\documentclass[runningheads,a4paper]{llncs}

\usepackage{url}
\usepackage{graphicx}
\usepackage{epstopdf}
\usepackage{paralist}
\usepackage{placeins}
\usepackage{cite}
\usepackage{wrapfig}
\usepackage{latexsym}
\usepackage{paralist}
\usepackage{xcolor}
\usepackage{xspace}
\usepackage{graphicx}
\usepackage{caption}
\usepackage{cleveref}
\usepackage{array}
\usepackage{booktabs}
\usepackage{xspace}
\usepackage{amsmath}
\usepackage{multirow}
\usepackage{wrapfig,lipsum,booktabs}

\usepackage{array}
\newcolumntype{P}[1]{>{\centering\arraybackslash}p{#1}}

\newcommand{\entity}[1]{\textit{#1}}
\newcommand{\tweet}[1]{\textit{`#1'}}
\newcommand{\todo}[1]{[\textbf{TODO:} \textit{\color{blue}#1}]}
\newcommand{\rephrase}[1]{\textit{\color{gray}#1}}
\newcommand{\changed}[1]{\textit{\color{brown}#1}}

\DeclareMathOperator*{\argmax}{\arg\!\max}

\renewcommand{\todo}[1]{}
\renewcommand{\rephrase}[1]{}

\renewcommand{\changed}[1]{#1}

\urldef{\mailsa}\path|{sujan, adarsh, amit, tkprasad}@knoesis.org, pnmendes@us.ibm.com|
\newcommand{\keywords}[1]{\par\addvspace\baselineskip
\noindent\keywordname\enspace\ignorespaces#1}

\begin{document}

\mainmatter  

\pdfinfo{
/Title (Implicit Entity Linking in Tweets)
/Author (Sujan Perera, Pablo Mendes, Adarsh Alex, Amit Sheth, Krishnaprasad Thirunarayan)
/Keywords (Implicit entities;Entity modeling;Entity linking;Knowledge graphs;Contextual knowledge;Dynamic knowledge)
/Subject (Implicit Entity Linking in Tweets)}

\title{Implicit Entity Linking in Tweets}

\titlerunning{Implicit Entity Linking in Tweets}

\author{Sujan Perera \and Pablo Mendes \and Adarsh Alex \and Amit Sheth \and Krishnaprasad Thirunarayan}
\authorrunning{Perera et al.}

\institute{Kno.e.sis Center, Wright State University\\
Dayton, Ohio, USA\\
IBM Research\\
San Jose, CA, USA\\
\mailsa\\}

\maketitle

\begin{abstract}
Over the years, Twitter has become one of the largest communication platforms providing key data to various applications such as brand monitoring, trend detection, among others.
Entity linking is one of the major tasks in natural language understanding from tweets and it associates entity mentions in text to corresponding entries in knowledge bases in order to provide unambiguous interpretation and additional context.
State-of-the-art techniques have focused on linking explicitly mentioned entities in tweets with reasonable success.
However, we argue that in addition to explicit mentions -- i.e. \tweet{The movie Gravity was more expensive than the mars orbiter mission} -- entities (movie Gravity) can also be mentioned implicitly -- i.e. \tweet{This new space movie is crazy. you must watch it!.}
This paper introduces the problem of implicit entity linking in tweets.
We propose an approach that models the entities by exploiting their factual and contextual knowledge.
We demonstrate how to use these models to perform implicit entity linking on a ground truth dataset with 397 tweets from two domains, namely, Movie and Book.
Specifically, we show: \begin{inparaenum}[1)]
\item the importance of linking implicit entities and its value addition to the standard entity linking task, and 
\item the importance of exploiting contextual knowledge associated with an entity for linking their implicit mentions. 
We also make the ground truth dataset publicly available to foster the research in this new research area.
\end{inparaenum}
\keywords{Implicit entities, Modeling entities, Entity linking, Knowledge graphs, Contextual knowledge, Dynamic knowledge}
\end{abstract}

\section{Introduction}
\label{sec:intro}
Data show that 350,000 tweets are generated per minute -- 500 million per day.\footnote{http://www.internetlivestats.com/twitter-statistics}
These tweets have become a valuable source of information for trend detection, event monitoring, and opinion mining applications.
Mining tweets poses unique challenges due to their short, noisy, context-dependent, and dynamic nature \cite{derczynski2015analysis}.

Entity linking in tweets has the potential to benefit all aforementioned applications.
The term `entity' in this paper refers to an unambiguous, terminal page in Wikipedia as in \cite{guo2013link}.
State-of-the-art entity linking solutions in tweets have mainly focused on explicitly mentioned entities  \cite{chang2014e2e}\cite{derczynski2015analysis}\cite{guo2013link}\cite{liu2013entity}.
However, we will show that entities may also be mentioned implicitly.
For example, consider the two tweets: \tweet{movie wasn't the story Veronica wrote. It was the story director hacksawed!!} and \tweet{`whew' the movie was the WORST example of that book. Neil Burger completely rewrote that whole story!.}
State-of-the-art entity linking systems may link the entity mention \entity{Veronica Roth} in the first tweet and \entity{Neil Burger} in the second tweet to their corresponding entities in a knowledge base.
However, they do not realize that both tweets have implicit mentions of the movie \entity{Divergent}. 
We term entities that are being implicitly mentioned as `implicit entities'.

Linking implicit entities in tweets is an important task that affects downstream analytics. 
If they were ignored, a sentiment analysis task wouldn't identify that the aforementioned tweets had negative sentiment towards the movie \entity{Divergent}. 
An trend detection application wouldn't detect that \entity{Oscar Pistorius} is trending if it was not able to identify the reference\footnote{We use the terms `entity reference' or `entity mention' interchangeably to signify the usage of a phrase that unambiguously evokes a given entity in a tweet.} to him in a tweet like \tweet{Kinda sad to hear about that South African runner kill his girlfriend} as he was referred to frequently with similar phrases in tweets. 


We hypothesize that implicit entities are a common occurrence. 
In \Cref{sec:stats}, we assess the prevalence of implicit versus explicit entities on a random sample of tweets.
This experiment shows that 21\% of the entities are mentioned implicitly in the Movie domain while it is 40\% in the Book domain.
Therefore, linking these entities will have significant impact on downstream applications.

The implicit entity linking problem (IEL) is notably different from explicit entity linking (EL) in several ways.
While we elaborate our findings in \Cref{sec:understanding}, we summarize them here.
Implicit entity mentions do not contain the entity name.
The absence of the entity name is filled by leveraging different characteristics of the entity -- the first tweet in the above examples uses the author of the book that the movie is based on while the second tweet uses its director to make a reference to the movie.
Furthermore, the context that helps to resolve the implicit entity mentions changes overtime -- the phrase `space movie' may refer to distinct movies at different time intervals.
These features of implicit entity mentions warrant a new approach to solve this problem.


In this paper, we propose an approach to the implicit entity linking problem in tweets that factors in the above features.
Twitter users often rely on sources of context outside the current tweet, assuming that there is some shared understanding between them and their audience, or temporal context in the form of recent events or recently-mentioned entities \cite{derczynski2015analysis}. 
This assumption allows them to constrain the message to 140 characters, yet make it understandable to the audience.
Our approach models entities by encoding this shared understanding by harnessing factual and contextual knowledge of entities to complement the context expressed in the tweet text.
The contextual knowledge captures temporally relevant topics and other entities associated with the entity of interest.


Our evaluation shows that the proposed approach achieves 61\% of accuracy in linking implicit entities. 
Furthermore, we show that IEL helps boost the overall accuracy of entity linking when combined with the EL. 
\todo{cant give a number for second point here, since it has done separately for each tool}
Our contributions are:
\begin{itemize}
\item We introduce the IEL problem on tweets, assess the significance of implicit entity mentions in a sample of tweets and describe their characteristics,
\item We propose and evaluate a model to capture and encode both factual and contextual  knowledge to perform implicit entity linking, and 
\item We create ground truth data sets and make them publicly available to foster research on this important new problem.
\end{itemize}

\section{Understanding Implicit Entity Mentions}
\label{sec:stats}
\label{sec:understanding}

In this section, we formally define the IEL problem w.r.t tweets and describe its main characteristics. 
We discuss the prevalence of implicit mentions on Twitter, the dynamicity of context associated with the entities, and the types of references-through-characteristics.

\begin{definition}{Implicit Entity}
is an entity mentioned in a tweet where its name is not present nor it is a synonym/alias/abbreviation of an entity name or a co-reference of an explicitly mentioned entity in the tweet.
\end{definition}

The explicit entity linking (EL) task can be defined as ``matching a textual entity mention, possibly identified by a named entity recognizer, to a KB entry, such as a Wikipedia page that is a canonical entry for that entity'' \cite{rao2013entity}. 
Therefore, the input for EL is a tuple $(s,c,t)$ where $s$ is the mention string (i.e. surface form) and $c$ is the entity type extracted by NER from text $t$. The output is an entity identifier $e$ such that $\argmax_{e} P(e|s,c,t)$.

We define the implicit entity linking for tweets as:
\begin{definition}{Implicit Entity Linking (IEL)}: 
given a tweet with an implicit entity mention of a particular type (e.g. Movie, Book) output the entity mentioned by the tweet w.r.t a given knowledge base.
\end{definition}

Therefore, the input for IEL is a tuple  $(c,t)$ where $c$ is the entity type and $t$ is the text (e.g. tweet) where the implicit entity occurred.
In contrast to EL 
the candidate set for an implicit entity mention is potentially much larger since it cannot be narrowed down based on the name of the entity.

\subsection{Prevalence}
\label{sec:prevalence}
To estimate the volume of implicit entity mentions on tweets, we performed a manual analysis on a sample set of tweets.
We focused on the domains of Movies and Books, since they offered an agreeable level of difficulty for human annotators.
We collected two random samples of tweets -- one for the Movie domain using the keywords `movie' and `film', and another for the Book domain using the keywords `book' and `novel'.
We subsequently annotated the tweets in the sample as `explicit', `implicit', and `NIL'  according to the following guidelines.

Consider three tweets in the Movie domain:
\begin{inparaenum}[1)]
\item \tweet{the movie trailer for 50 shades of grey looks really good,} 
\item \tweet{ISRO sends probe to Mars for less money than it takes Hollywood to make a movie about it,} and
\item \tweet{How the hell is every movie the \#1 movie in America?.}
\end{inparaenum}
The first tweet has a mention of the movie \entity{Fifty Shades of Grey}, hence it is annotated as `explicit.' 
The second tweet is annotated as `implicit' since it has an implicit reference to the movie \entity{Gravity.}  
The third tweet does not refer to any movie, hence it is annotated as `NIL.'
The tweets that have both explicit and implicit movie mentions are annotated with both labels.

This annotation exercise produced 416 and 114 tweets for the `explicit' and `implicit' categories respectively. 
This means 21\% of the tweets with mentions of movies are implicit references.
In other words, this experiment showed that roughly for every four tweets with explicit mentions of entities, there is a tweet with an implicit mention in the Movie domain.
A similarly constructed experiment in the Book domain found that roughly for every 5 tweets with explicit entity mentions there are two tweets with implicit mentions.

\subsection{Characteristics}
\subsubsection{Dynamic Context}
When a human annotator is trying to resolve an implicit entity mention, they often rely on domain knowledge outside the tweet text -- for instance, they know which were the latest movies, or which actors starred in different movies, etc. 
The relevant domain knowledge may change dynamically.

Often, one phrase may refer to distinct entities at different time intervals.
For instance, the phrase `space movie' could refer to the movie \entity{Gravity} in Fall 2013 while the same phrase in fall 2015 would likely refer to the movie \entity{The Martian}.
On the flip side, the most salient characteristics of the movies may change over time, and so will the phrases used to refer to them.
The movie \entity{Furious 7} was frequently referred to with phrase `Paul Walker's last movie' in November 2014. This was due to the actor's passing around that time. 
However, after the movie release in April 2015 the same entity was often mentioned through the phrase `fastest film to reach the \$1 billion.' 

\subsubsection{Types of references-through-characteristics}
We observed that individuals resort to a diverse set of entity characteristics to make implicit references.
For example, consider the following implicit references to the movie \entity{`Boyhood'}\footnote{We show only the fragments of the tweets that indicate a mention of an entity.}: 
\begin{inparaenum}[1)]
\item `Richard Linklater movie,' 
\item `Ellar Coltrane on his 12-year movie role,' 
\item `12-year long movie shoot,' 
\item `latest movie shot in my city Houston,' and
\item `Mason Evan's childhood movie.'
\end{inparaenum}
The first two tweet fragments \todo{are they really tweets or are they fragments?} refer to the movie through its director and actor, the third tweet fragment uses a distinctive feature (it was shot over a long period), the fourth example uses the shooting location of the movie, and last one refers to it with a character in the movie. \todo{PABLO: can we get a more comprehensive list of ways in which people refer to the movies in your dataset?}
\todo{We don't have to, but this is an opportunity for a table: X mentions using characteristics, Y using temporal expressions, etc.}

\section{Related Work}
\label{sec:related}

Entity linking in tweets has recently gained attention in academia and industry alike.
The literature on entity linking in tweets can be categorized as `word-level entity linking' and `whole-tweet entity linking' \cite{derczynski2015analysis}.
While the former task is focused on resolving the entity mentions in a tweet, the latter task is focused on deriving the topic of the tweet.
The topic may be derived based on the explicit and implicit entity mentions in the tweets.
For instance, the tweet \tweet{Texas Town Pushes for Cannabis Legalization to Combat Cartel Traffic} has an explicit mention of entity \entity{Cannabis Legalization} and an implicit mention of entity \entity{El Paso} city. 
The topic of the tweet would be `Cannabis Legalization in El Paso.'
Hence, it is worth noting that the work on deriving topics is neither comparable to explicit nor to implicit entity linking since they are extracting the topic of the tweet text, rather than actual mentions of an entity in the tweet \cite{derczynski2015analysis}.
We have not found literature that focuses on implicit entity linking in tweets. 
In this section, we will survey the literature on both word-level and whole-tweet entity linking and explain why techniques and features used by such solutions may not be applicable to implicit entity linking, hence it deserves special attention. 

Meij, et al. \cite{meij2012adding} derives the topics of a tweet.
They extract features from the tweet and the Wikipedia pages of entities, and apply machine learning algorithms to derive the topic.
We found that this work has focused on deriving topic using explicit entities as their evaluation dataset contains only 16 tweets whose label of the manually annotated topic is not present in the tweet text (i.e. not a string match).
Nevertheless, they are found to be either synonyms or related entities to the explicit entities in the tweets and not implicit entity mentions (e.g. New York and Big Apple, Almighty and God, stuffy nose and Rhinitis).

The word-level tweet linking has two main steps: \begin{inparaenum}[1)]
\item candidate selection, and 
\item disambiguation. 
\end{inparaenum}
The word-level entity linking has been studied extensively for text like Wikipedia and News   \cite{cucerzan2007large}\cite{dredze2010entity}\cite{hoffart2011robust}\cite{mendes2011dbpedia}\cite{milne2008learning}; however, these approaches have  proved to be ineffective on short and noisy text like tweets \cite{derczynski2015analysis}\cite{ferragina2010tagme}.
Here we will discuss the approaches taken to solve this problem in tweets.
The first step was performed by matching the word sequences of the tweet to the page titles and the anchor texts in Wikipedia and consider all matching pages and pages redirected by matching anchor texts to be candidates \cite{chang2014e2e}\cite{ferragina2010tagme}\cite{guo2013link}.
The second step was performed by optimizing the relatedness calculated among the candidate entities \cite{ferragina2010tagme}\cite{liu2013entity} or based on the threshold defined over measures that calculate the similarity between the entity mention and entity representation \cite{chang2014e2e}, or by applying structural learning techniques \cite{guo2013link}.
An approach to solve the implicit entity linking in tweets has to take a fresh perspective since by definition it has neither anchor text nor the page title present in the tweet.

Our previous work on implicit entity linking has dealt with clinical entities in electronic medical records \cite{pereraimplicit}.
The main challenge in the medical setting resides in the heterogeneous usage of language by individuals mentioning entities implicitly including their negated mentions. 
Our approach focused on modeling the entities by using their static definitions and exploiting WordNet as the knowledge base to account for heterogeneity in language usage.
The task of implicit entity linking in tweets is different from that of clinical text since, in addition to the differences in nature of the text, the heterogeneity arise with the usage of different characteristics of entities and their association with dynamic  context as discussed in \Cref{sec:understanding}.
Hence, the static descriptions/definitions of entities fall short in linking implicit entities as we will demonstrate in our evaluation.

\section{Linking Implicit Entities in Tweets}
\label{sec:definition}

Our approach models entities with factual knowledge and contextual knowledge by leveraging existing knowledge bases and relevant tweets.
The entity models are integrated to create an entity model network (EMN) for each domain of interest to reflect the topical relationships among domain entities\footnote{The term `domain entity' in this work refers to movies in the Movie domain and books in the Book domain.} at time $t$.

The first step of creating the EMN is to identify domain entities that are relevant at time $t$.
This can be done, for instance, by running an off-the-shelf entity linking system over a corpus of recent tweets and identify the mentioned entities. 
The idea is that if an entity is relevant at time $t$, it will likely to be mentioned explicitly by tweets around that time. 
Even though an automatic annotation approach may not be perfect, it will identify at least one occurrence of explicit mention of entities within the corpus.
This is sufficient as one appearance of an entity in the corpus qualifies it to be included in the EMN.
We start building EMN by creating entity models for identified entities for time $t$.

\subsection{Entity Model Creation}
\label{sec:model}
\begin{figure*}
\includegraphics [scale=.4]{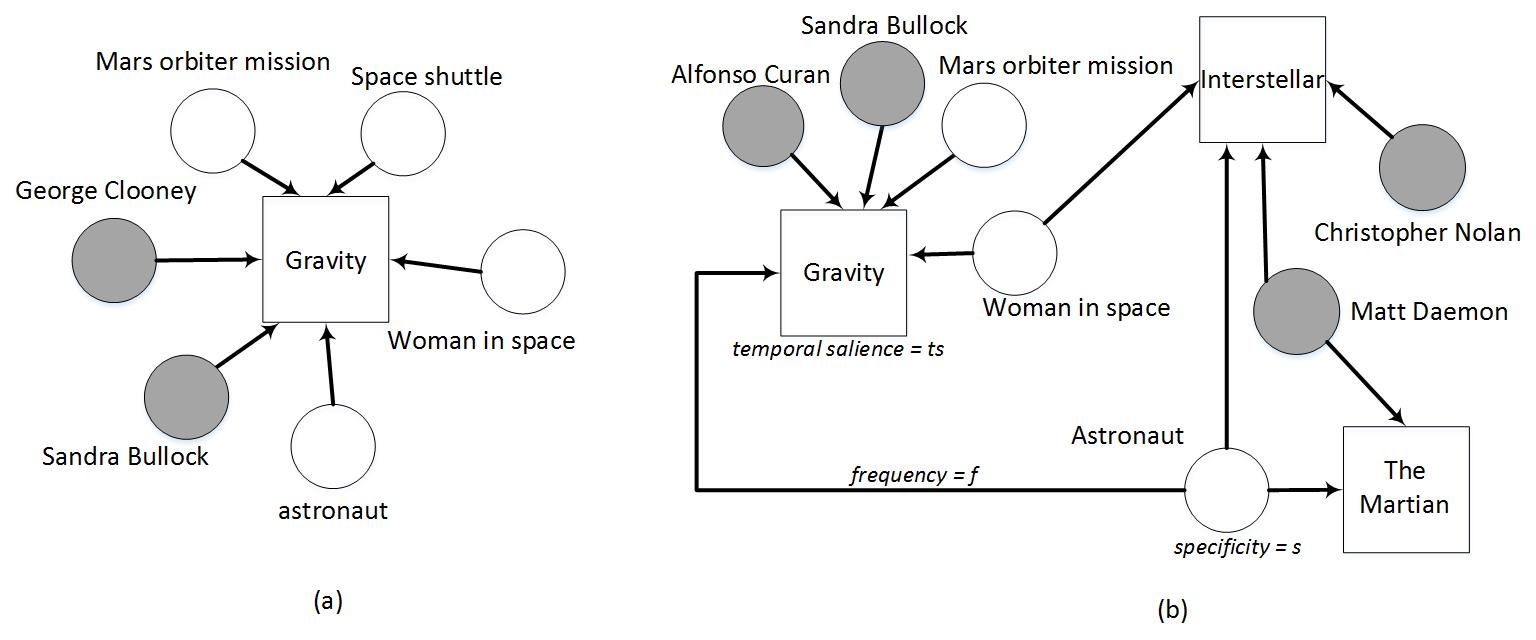}
\caption{Entity Model Network. The rectangles depict entities, shaded circles and plain circles show the clues generated with factual and contextual knowledge respectively. The properties of nodes and edges are shown for one connection in (b).}
\label{figure:EMN}
\end{figure*}

Consider two tweets about the movie \entity{Gravity}: \tweet{New Sandra Bullock astronaut lost in space movie looks absolutely terrifying,} and \tweet{ISRO sends probe to Mars for less money than it takes Hollywood to send a woman to space.}
The first tweet has a mention of its actress and that, along with other terms, helps to resolve its mention to the movie \entity{Gravity}. 
This kind of \textbf{factual knowledge} (e.g. to relate actors and movies) can be extracted from a knowledge base.
The second tweet does not have a mention of any entity associated with the movie \entity{Gravity}; hence, the factual knowledge falls short in identifying its implicit mention.
However, contemporary tweets with explicit mentions of movie \entity{Gravity} often use phrases like `ISRO', `woman to space', `less money' which helps to link the implicit mention of \entity{Gravity} in the second tweet. 
We refer to these as \textbf{contextual knowledge}.
\newcommand{\popularity}{temporal salience\xspace}
The entity model also contains an estimate of \textbf{\popularity} for each entity and it is computed by counting the corresponding Wikipedia page views within the last 30 days w.r.t time $t$.
The entity model consists of phrases generated using the knowledge components and its \popularity.
\Cref{figure:EMN} (a) shows the fragment of the model generated for the movie \entity{Gravity}.

\textbf{Acquiring factual knowledge:}
Factual knowledge of an entity can be acquired from existing knowledge bases and Linked Open Data.
We have used DBpedia \cite{lehmann2014dbpedia} as our knowledge base due to its wide coverage of domains and up-to-date knowledge.
For a given entity $e$ we retrieved triples where $e$ appears as subject or object.
However, for a given entity type not all relationships are important in modeling its entities.
For example, a movie has relationships `director' and `starring' as well as `billed' and `license.' 
The former two relationships are more important when describing a movie than the latter.
We capture this intuition by ranking the relationships based on their joint probability value with the given entity type as follows.

\begin{equation}
\label{eq:relrank}
P(r, T)=\frac{\textit{number of triples of r with instances of T}}{\textit{total number of triples of r}},
\end{equation}
where T is the entity type (e.g. Movie) and $r$ is the relationship. 
The instances of a given entity type can be obtained from DBpedia via `rdf:type' relationship.

The triples of entity $e$ of type $T$ with one of the top \textit{m} relationships are selected to build the entity model of $e$.
We collect the `rdfs:label' value of entities connected to $e$ in these triples as the factual knowledge of entity $e$.
In addition to the top \textit{m} relationships, we also consider the value of `rdfs:comment' relationship of $e$ to build the entity model.
`rdfs:comment' gives a textual description of an entity that oftentimes complements the knowledge captured by the triples.

\textbf{Acquiring contextual knowledge:}
The contextual knowledge can be extracted from contemporary tweets that explicitly mention the entity.
We use rdfs:label of the entity in DBpedia along with its type as the keyword to collect the 1000 most recent tweets for that entity.
For example, we used the phrases `gravity movie' and `gravity film' to collect tweets for  movie \entity{Gravity}; this will minimize the tweets with other meanings of the term `gravity' in collected tweets.

\textbf{Model Creation:}
The entity model consists of weighted phrases and unigrams generated using the acquired knowledge and the \popularity of the entity.
Firstly, the collected tweets needs to be cleaned before they can be used to create the entity model.
We remove the punctuations, and emoticons, and normalizes the numbers to the pseudo-string `NUMBER' in tweets.
The hashtags and mentions that were written in camel case style were retained after decomposing (@VeronicaRoth $\rightarrow$ Veronica Roth, \#MarkWahlberg $\rightarrow$ Mark Wahlberg) and others were retained by removing `\#' and `@' symbols. \todo{beware of twitter terms of service https://about.twitter.com/company/display-requirements}

We use Wikipedia anchor texts and page titles to identify meaningful phrases in the acquired knowledge. 
We chunk the text in acquired knowledge \changed{(i.e. factual knowledge and cleaned tweets)} into n-grams (n=2, 3, 4) and the n-grams present as anchor text or as page titles are added to the entity model as phrases.
However, Twitter users do not always use complete phrases; consider the reference to actress \entity{Sandra Bullock} in the tweet: \tweet{It's hard for me to imagine movie stars as astronauts, but the movie looks great! and who doesn't like \textbf{Sandra}.}
Therefore, the entity model should also contain the portions of the phrases, hence we include unigrams excluding stop words to fulfill this requirement.

An entity model consists of phrases as well as unigrams (collectively referred as clues), and we stored it as a graph as shown in the \Cref{figure:EMN}(a). 

\subsection{Entity Model Network (EMN) Creation}
The entity models created for all the entities are integrated via common clues to generate the EMN as shown in \Cref{figure:EMN}(b).
The specificity of the clue node $c_j$ is inspired by the inverse document frequency measure and is calculated as $\log \frac{|N|}{|N_{c_j}|}$ where $|N|$ is the number of entity nodes in the EMN and $|N_{c_j}|$ is the number of adjacent nodes to $c_j$.
The `frequency' property value of the edge between clue node $c_j$ and entity node $e_i$ is calculated as the total number of times that the value of `clue name' of node $c_j$ is present in tweets collected for entity $e_i$.

Formally, an entity model network (EMN) is defined as a property graph $G_{EMN}= (V_e, V_c, E, \mu)$, where $V_e$ and $V_c$ represent the vertices of two types, $E$ represents the edges, and $\mu$ represents the property map. The edges are directed (i.e. $E \subseteq (V_c \text{ X }  V_e)$), and $\mu$ maps the properties of vertices and edges as keys to values (i.e. $\mu : (V_e \cup V_c \cup E) X R \rightarrow S$),
where $R$ is a set of keys and $S$ denotes values.
$V_e$ represents the entities and has the properties `name' and `\popularity' as keys and their values as key/value pairs.
$V_c$ represents the clues and has the properties `clue name' and `specificity' as keys and their values as key/value pairs.
The edges in the graph has the property `frequency' and its value as key/value pair.

\subsection{Linking Implicit Entities with the EMN}

To understand how to use the EMN to perform implicit entity linking for a given tweet, it is useful to divide the task into two steps: \begin{inparaenum}[1)]
\item candidate selection and ranking, and 
\item disambiguation.
\end{inparaenum}

\subsubsection{Candidate Selection and Ranking}
The objective of the candidate selection and ranking step is to prune the search space so that disambiguation step does not have to evaluate all entities in EMN as candidates.
In EL this is usually done by looking for candidates for a given surface form.
In IEL, we took a different approach.
The input tweet goes through the cleaning step as described in \Cref{sec:model}. 
Then we identify the phrases in the tweet using Wikipedia anchor text and page titles, and the terms that are not qualified as phrases are considered as unigrams.
\todo{tweet clues or entity clues or context clues? I like context clues better.}
We refer to both phrases and unigrams extracted from tweets as `tweet clues' and denotes it as $\mathcal{C}_t$.
The candidate selection step takes these tweet clues and match with clue nodes in EMN, the entities that have at least one edge from matching clues are selected as the initial set of candidates.

Formally, given a set of tweet clues $\mathcal{C}_t$, the initial candidate entity set $\mathcal{E_{IC}} = \{e_i | (c_j, e_i) \in E \textit{ and } c_j \in \mathcal{C}_t\}$.

The entities in the initial candidate set are scored based on the strength of evidences. The strength of evidences for entity $e_i \in \mathcal{E_{IC}}$ ($SC_{e_i}$) is calculated as:

$SC_{e_i}=\sum_{{c_j}\in \mathcal{C}_t} \textit{specificity of }c_j*\textit{frequency of edge } (c_j, e_i)$

The top k candidates based on these scores (denoted as $E_c$) are considered for disambiguation step.

\subsubsection{Disambiguation}
The objective of the disambiguation step is to sort the selected candidate entities such that the implicitly mentioned entity in a given tweet is at the top position of the ranked list.
This is accomplished through a machine learned-ranking model based on the pairwise approach: all pairs of selected candidate entities (along with a feature set) are taken as input, and the model approximates the ranking as a classification problem that tells which of the entities in the pair is better than the other.

The feature set of a candidate entity consists of its similarity to the tweet and its \popularity w.r.t \popularity of other candidate entities.

The similarity between the candidate entity and the tweet is calculated via their vector representations.
The vector representation of the candidate entity $e_i$ is obtained via its incoming connections from other nodes.
It is denoted as $e_{i_v}$ and defined as $e_{i_v}=<v_1, v_2, ..., v_n>$ where $v_j=\textit{specificity of } c_j*\textit{frequency of edge } (c_j, e_i)$ for all $(c_j, e_i) \in E$.
The vector representation of the tweet is created using tweet clues.
The similarity between the candidate entity and the tweet is calculated by the cosine similarity of these vectors.
\todo{flow is confusing}

The \popularity of the candidate entity $e_i$ is normalized w.r.t the \popularity of other candidate entities in $E_c$ as:
\begin{equation}
\frac{\textit{\popularity of } e_i}{\sum_{e\in E_c} \textit{\popularity of } e}
\end{equation}

We trained a SVM$^{rank}$ model to solve the ranking problem. 
\changed{We used linear kernel, 0.01 as the trade-off between training error and margin, and total number of swapped pairs summed over all queries as the loss function.}
SVM$^{rank}$ shown to perform well in similar ranking problems, specifically it is able to provide best performance in ranking the top concept \cite{meij2012adding} which suits the characteristics of our problem.

\section{Evaluation}
\label{sec:eval}

We evaluated the implicit entity linking performance and its value to the explicit entity linking task in two domains, namely, Movie and Book.  
There is no standard dataset available to evaluate this task.
Hence, we have created datasets for the two domains.
We focus on answering three questions in our evaluation.

\begin{itemize}
\item How effective is the proposed approach in linking implicit entities?
\item How important is the contextual knowledge in linking implicit entities?
\item What is the value added by linking implicit entities?
\end{itemize}

\subsection{Dataset Preparation}

In order to prepare datasets for evaluation, we collected tweets with `movie' and `film' as keywords for the Movie domain and `book' and `novel' as keywords for the Book domain.
\changed{This dataset was collected during August 2014 and it was independent of the dataset described in \Cref{sec:prevalence}.}
The collected tweets were manually annotated by two individuals with `explicit', `implicit', and `NIL' labels for Movies and Books following the guidelines described in \Cref{sec:prevalence}.
We included annotated tweets to the evaluation dataset that was agreed upon by both annotators.
\changed{\Cref{table:datastat} shows the important characteristics of the annotated dataset and it is available in our project page at \url{https://goo.gl/jrwpeo}.} 

\begin{table}
\begin{center}
\begin{tabular}{ p{1.8cm}  p{2.0cm}  p{2cm}  p{3.5cm}  p{2cm} } 
\hline
 \textbf{Domain} & \textbf{Annotation} & \textbf{Tweets} & \textbf{Entities} & \textbf{Avg. length}\\ \hline
\hline
\multirow{3}{4em}{Movies} & explicit & 391 & 107 & 16.5 words \\ 
& implicit & 207 & 54 & 18 words \\ 
& NIL & 117 & 0 & 16.4 words \\ 
 \hline 
\multirow{3}{4em}{Books} & explicit & 200 & 24 & 18.5 words \\ 
& implicit & 190 & 53 & 18.5 words \\ 
& NIL & 70 & 0 & 17.5 words \\
\hline
\end{tabular}
\caption{Evaluation Dataset Statistics. Describes, per domain, the total number of tweets per annotation type (explicit, implicit, NIL),  number of distinct entities annotated, and average tweet length.}
\label{table:datastat}
\end{center}
\end{table}

To perform IEL on the aforementioned evaluation dataset, we created the EMN for 31st of July 2014.
We collected most recent 15,000 tweets to the date 31st of July 2014 for each domain using its type labels as the keywords (e.g. `movie' and `film' for movie domain) and applied a simple spotting mechanism to identify entity mentions.
The spotting mechanism collects the labels of the domain entities from DBpedia (i.e. rdfs:label value of the instances of type Film) and then it checks for their presence in collected dataset for the domain.
If the label is found within the tweets, we add that entity to the EMN.
We collected 1000 most recent tweets that explicitly mention the identified entities to generate their contextual knowledge, extracted the factual knowledge from DBpedia version created with May 2014 Wikipedia dumps, and obtained page hit counts of Wikipedia pages for the month of July 2014.
\changed{We varied the number of top \textit{m} relationships ranked according to equation  \ref{eq:relrank} to extract the factual knowledge for modeling entities. 
The best results were obtained when $m=10$ and we observed dramatic decrease in the accuracy in linking movie entities when $m>15$, potentially due to inclusion of relatively irrelevant knowledge. 
Hence, the factual knowledge component consists of knowledge extracted with top 15 relationships.}
These collected resources are used to create entity models for each entity and the EMN as described in \Cref{sec:definition}.
The created EMNs for the Movies and Books domains had 617 entities and 102 entities respectively.

\begin{table}
\begin{center}
\begin{tabular}{P{2cm}  P{5cm}  P{4.5cm}}
    \hline
    \textbf{Domain} & \textbf{Candidate Selection Recall} & \textbf{Disambiguation Accuracy} \\ \hline
   Movie & 90.33 & 60.97 \\ \hline
   Book & 94.73 & 61.05 \\ \hline 
\end{tabular}
\caption{Implicit Entity Linking Performance}
\label{table:results}
\end{center}
\end{table}

\subsection{Implicit Entity Linking Evaluation}
\label{eval1}
This section evaluates the implicit entity linking task in isolation.
We show the results on both candidate selection and ranking, and disambiguation steps.
\changed{The candidate selection and ranking is evaluated as the proportion of tweets that had the correct entity within the top k selected candidates for that tweet (denoted as Candidate Selection Recall).
We experimented by varying k between 5 and 35 and found that results improves as we increase k and comes to near plateau after k=20.
We demonstrate the results for k=25 and interested readers can find detailed results on our project page.}
The disambiguation step is evaluated with 5-fold cross validation and report the results as proportion of the tweets in evaluation dataset that had correct annotation at the top position. 
\Cref{table:results} shows the results of this step on both domains.

\paragraph{Qualitative Error Analysis}
The error analysis on implicit entity linking shows that errors are fourfold: \begin{inparaenum}[1)]
\item errors due to lack of contextual knowledge of the entity, 
\item errors due to novel entities,
\item errors due to cold start of entities and topics, and
\item errors due to multiple implicit entities in the same tweet.
\end{inparaenum}
\changed{\Cref{table:qualitative} shows an example tweet for each error type.}

\begin{table}
\begin{center}
\begin{tabular}{ p{1cm} p{11cm}} 
\hline
 \textbf{Error} & \textbf{Tweet} \\ \hline
\hline
1 & \tweet{That Movie Where Shailene Woodley Has Her First Nude Scene? The Trailer Is RIGHT HERE!: No one can say Shailene Woodley isn't brave!} \\ 
 \hline 
2 & \tweet{``hey, what's wrawng widdis goose?" RT @TIME: Mark Wahlberg could be starring in a movie about the BP oil spill http://ti.me/1oZh55V} \\  \hline 
3 & \tweet{Video: George R.R. Martin's Children's Book Gets Re-release http://bit.ly/1qNNH5r} \\  \hline 
4 & \tweet{That moment when you realize that hazel grace and Augustus are brother and sister in one movie and in love battling cancer..} \\ 
\hline
\end{tabular}
\caption{Example Tweet for Each Error Type.}
\label{table:qualitative}
\end{center}
\end{table}

\changed{The first tweet in the table is annotated with movie \entity{White Bird in a Blizzard}.
There were only 46 tweets for this movie.
Hence, the contextual knowledge component of the entity model did not provide strong evidences in disambiguation step.
The second tweet is annotated with movie \entity{Deepwater Horizon}. 
The Wikipedia page for this movie was created on September 2014, hence it was not available to EMN.
This is known as emerging entity discovery problem and requires separate attention as in the explicit entity linking \cite{hoffart2014discovering}.
A few entities and topics emerged among Twitter users only after 31 July 2014.
These entities and topics were not present in our EMN.
One of them is the republication of George Martin's book \entity{The Ice Dragon} which is emerged in early August 2014 resulting tweets about the book.
One such tweet is showed in the third row in \Cref{table:qualitative}.
Both the entity and the topic were not known to the EMN, hence it couldn't link tweet to the book \entity{The Ice Dragon}.
This problem can be solved by implementing an evolution mechanism for EMN. 
Lastly, a couple of tweets in the dataset had two implicit entities. 
One such tweet is shown in last row of the table which is annotated with movies \entity{The Fault in Our Stars} and \entity{Divergent}.
Since our method links only one entity per tweet, we had the choice to either remove the tweet with two mentions or add it once for each mention. 
We did the latter to preserve the characteristics of the tweets.
However, not surprisingly, both tweets were annotated with the same movie (\entity{The Fault in Our Stars}) resulting an incorrect annotation.
Although this is a limitation of our approach, this phenomenon is not a frequent occurrence as the dataset had only 6 (=1.5\%) tweets with two implicit entity mentions.}

\subsection{Importance of Contextual Knowledge}

One of the major components in our entity model is the contextual knowledge of the entity.
This section evaluates its contribution to the proposed implicit entity linking solution by comparing the results obtained by EMN created with contextual knowledge and EMN created without contextual knowledge.
As in \Cref{eval1}, we evaluated this for both candidate selection and ranking, and disambiguation steps.

\begin{table}
\begin{center}
\begin{tabular}{ p{5.5cm} p{1.5cm}  p{2.3cm}  p{1.5cm} } 
\hline
 \textbf{Step} & \textbf{Domain} & \textbf{Without ctx} & \textbf{With ctx} \\ \hline
\hline
\multirow{2}{*} {Candidate Selection Recall} & Movie & 77.29\% & 90.33\% \\ 
& Book & 76.84\% & 94.73\% \\ 
 \hline 
\multirow{2}{*} {Disambiguation} & Movie & 51.7\% & 60.97\% \\ 
& Book & 50.00\% & 61.05\% \\ 
\hline
\end{tabular}
\caption{Contribution of the Contextual Knowledge Component of Entity Model.}
\label{table:FC}
\end{center}
\end{table}

\Cref{table:FC} shows the results of this experiment.
As shown in \Cref{table:FC}, the contextual knowledge contributes to both candidate selection and ranking, and disambiguation steps on both domains.
Quantitatively, the recall value of the first step increased by 14\% and 19\% while accuracy of second step increased by 15\% and 18\% for movies and books, respectively.
This is due to the fact that people do not necessarily use associated entities when referring to entities implicitly, but rely on other clues.
For example, consider tweets \tweet{``My name was Salmon, like the fish; first name, Susie." Great book!} and \tweet{`2 actors playing brother and sister then plot twist new movie, but they have cancer and love each other.'}
The first tweet has implicit mention of book \entity{The Lovely Bones} and second tweet has implicit mentions of movies \entity{The Fault in Our Stars} and \entity{Divergent.}
However, none of them contain any associated entities (e.g. author, publisher, actors, directors) to the book or the movie, hence, the factual knowledge component in the entity model fell short in these tweets.
The contextual knowledge component fills in the gaps since it can build the association between the clues indicated in tweets and the respective entities.

\subsection{Value Addition to Standard Entity Linking Task}

The real world datasets collected via keywords contain explicit and implicit entity mentions as well as tweets with no entity mentions.
This experiment assess the impact of implicit entity linking in such datasets.
To create a dataset for this experiment we followed the following steps:

\begin{itemize}
\item Select 40\% of tweets from dataset with implicit entities as the test dataset. We use the rest to train the ranking model.
\item Mix the selected test dataset with tweets with explicit entity mentions by preserving the explicit:implicit ratio. This ratio is 4:1 in the Movie domain and 5:2 in Book domain.
\item Add 25\% of tweets that has no mention of entity to account for NIL mentions.
\end{itemize}

\begin{table}
\begin{center}
\begin{tabular}{  p{2.5cm} p{1.5cm}  p{1.5cm}  p{1.5cm}  p{1.5cm}  p{1.5cm} p{1.5cm} } 
\hline
& \multicolumn{2}{ c }{\textbf{DBpedia Spotlight}} & \multicolumn{2}{ c }{\textbf{TagMe}} & \multicolumn{2}{ c }{\textbf{Zemanta}} \\ \hline
\hline
  & Movies & Books & Movies & Books & Movies & Books \\ \hline
\hline
F1 (EL) & 0.18 & 0.44 & 0.24 & 0.19 & 0.32 & 0.17  \\ 
F1 (EL+IEL) & 0.34 & 0.54 & 0.30 & 0.34 & 0.39 & 0.37  \\
\hline
\end{tabular}
\caption{EL and IEL combined Performance}
\label{table:aggregated}
\end{center}
\end{table}

\changed{The experiment setup used three well-known entity linking solutions with the following configurations: \begin{inparaenum}[1)]
\item DBpedia Spotlight \cite{isem2013daiber} (confidence=0.5, support=20),
\item TagMe \cite{ferragina2010tagme}  ($\rho$=0.5), and
\item Zemanta.\footnote{http://www.zemanta.com}
\end{inparaenum}}
It first annotates the prepared Movie dataset using DBpedia Spotlight.
The tweets that are not annotated with movies of the output are sent to our proposed solution assuming they have implicit entity mentions.
The same exercise is repeated for TagMe and Zemanta.
Then we conducted this experiment for Book dataset.

\Cref{table:aggregated} shows the results of this experiment using F1 measure.
\changed{The precision (P) and recall (R) values for F1 measure is calculated as follows:
\[
  P = \frac{\textit{c}}{\textit{total tweets annotated with entity}} \qquad\text{and}\qquad R = \frac{\textit{c}}{\textit{total tweets with entity}}
\]
where $c$ is the total number of correctly annotated tweets with an entity.}
These results demonstrate the value of adding IEL as a post-step to EL.

\subsection{Discussion}
\label{sec:discuss}

While we collected tweets based on a limited set of keywords, we do not depend on these keywords to link the implicit entities.
It is merely used as technique to collect tweets to create evaluation dataset.
Our approach can be applied to link the implicit entity mentions of a given type in the absence of these keywords.

We showed the value of contextual knowledge in implicit entity linking. 
We believe that it can play a similar role in the disambiguation step of the explicit entity linking task.


The requirement to evolve the EMN is observed with the first experiment.
We have identified the two events that can change the EMN over time:
\begin{inparaenum}[1)]
\item A new entity becomes popular and people start to tweet about it or the popularity of an existing entity fades away, and
\item A new topic of interest emerges for an existing entity or with the introduction of a new entity, or the popularity of the existing topic fades away.
\end{inparaenum}
In future, we will implement the operators that will keep EMN up-to-date by continuously collecting the tweets and injecting derived knowledge from the them, as well as from DBpedia.

\section{Conclusion and Future Work}
\label{sec:conclusion}

We introduced the problem of implicit entity linking in tweets and studied its prevalence and characteristics.
We proposed a solution that models the entities with their factual and contextual knowledge and demonstrated that these models are capable of linking implicit entities with higher accuracy than state-of-the-art entity linking approaches.
In the future, we will extend our model to account for NIL mentions and expand our evaluation to more domains and larger datasets.
Another interesting topic to study is the dynamics of the implicit entity references over time.


\bibliographystyle{splncs03}

\end{document}